\documentclass[
]{template2/ceurart}

\newcommand{\triple}[1]{{\small \textsc{#1}}}

\usepackage{listings}
\begin{document}

\copyrightyear{2025}
\copyrightclause{Copyright for this paper by its authors.
  Use permitted under Creative Commons License Attribution 4.0
  International (CC BY 4.0).}

\conference{TRUST-AI: The European Workshop on Trustworthy AI. Organized as part of the European Conference of Artificial Intelligence - ECAI 2025. October 2025, Bologna, Italy.}

\title{Bridging the AI Trustworthiness Gap between Functions and Norms (Position Paper)}

\author[1,2]{Daan {Di Scala}}[%
orcid=0000-0003-1548-6675,
email=daan.discala@tno.nl
]
\cormark[1]
\author[1]{Sophie Lathouwers}[%
orcid=0000-0002-7544-447X,
email=sophie.lathouwers@tno.nl,
]

\author[1]{Michael {van Bekkum}}[%
orcid=0009-0007-3009-254X,
email=michael.vanbekkum@tno.nl,
]

\address[1]{TNO Netherlands Organisation for Applied Scientific Research, Data Science Department}
\address[2]{Utrecht University, Department of Information and Computing Sciences}

\cortext[1]{Corresponding author.}

\begin{abstract}
Trustworthy Artificial Intelligence (TAI) is gaining traction due to regulations and functional benefits.
While Functional TAI (FTAI) focuses on how to implement trustworthy systems, Normative TAI (NTAI) focuses on regulations that need to be enforced.
However, gaps between FTAI and NTAI remain, making it difficult to assess trustworthiness of AI systems.
We argue that a bridge is needed, specifically by introducing a conceptual language which can match FTAI and NTAI. 
Such a semantic language can assist developers as a framework to assess AI systems in terms of trustworthiness. 
It can also help stakeholders translate norms and regulations into concrete implementation steps for their systems.
In this position paper, we describe the current state-of-the-art and identify the gap between FTAI and NTAI. 
We will discuss starting points for developing a semantic language and the envisioned effects of it. Finally, we provide key considerations and discuss future actions towards assessment of TAI.
\end{abstract}


\maketitle

\section{Introduction}

Trustworthy Artificial Intelligence (TAI) is increasingly recognised as essential for both development and deployment of AI systems to create trust and confidence with stakeholders and end users.
Various approaches emerge to make AI more trustworthy, each reflecting different perspectives.
The functional perspective focuses on creating AI systems with sufficient technical reliability and safety \cite{ahmad2022requirementsengineeringartificialintelligence}.
From a normative and legal point of view, TAI involves adherence to rules, standards, and compliance frameworks such as ISO/IEC 42001, ISO/IEC 23894 and the EU AI Act \cite{union2021proposal, benraouane2024ai, simonetta2024iso}.
Socially, the emphasis of trustworthiness lies on the broader societal impact of AI, which addresses issues such as power, ethics, privacy, inclusivity, and the systems' perceived benevolence \cite{hagerty2019global, marko2025examining, memarian2023fairness, novozhilova2024more}.

While many approaches have been proposed towards TAI, there is a clear lack of overlap and integration between functional and normative approaches.
As the AI Act focuses on norms for high-risk systems, many 
 existing initiatives provide risk assessment frameworks to check compliance \cite{ai-act-compliance-checker, ai-act-implementation-tool, ai-impact-assessment-nl, altai}. 

For example, the AI Risk Ontology (AIRO) \cite{golpayegani2022airo} has been proposed to determine which systems are high-risk and to document related risk information.
Although useful, these risk-based approaches focus on application categories of AI systems and remain largely disconnected from system design choices and concrete implementation steps. 

The past has shown that it can be difficult for developers to build systems that comply with regulations such as the General Data Protection Regulation (GDPR) \cite{sirur2018we}. Research has shown that this is in part because developers have trouble relating normative requirements to technical implementations \cite{senarath2018developers, alhazmi2021m}. Therefore, it was recommended to accompany the GDPR law with techniques to use for implementing each principle. We expect regulations on TAI such as the AI Act to face similar problems, as they do not provide accompanying techniques and guidelines for developers to use.

To bridge the gap between norms and regulations on one side and functional requirements on the other, we see the need for a mapping that 1) supports technical implementations in compliance with regulations 
and 2) translates legal standards into system requirements for developers. 
Without such a mapping, there is a risk that abstract or high-level trustworthiness principles remain disconnected from technical decisions during AI development and functional assessment of AI systems on TAI compliance proves elusive.

Therefore, we propose that a standardised semantic framework should be developed that relates functional properties of AI systems to key trustworthiness requirements. 
As we believe that the envisioned semantic framework should build on existing frameworks, we explore related work that can serve as a starting point. We include both normative frameworks and ways to systematically explore system designs. We then provide an initial design and conclude with key considerations and directions for future work.

\section{Towards a Bridging Semantic Framework}
We now describe key starting points to consider for a semantic framework that bridges the gap between functional and normative AI requirements. For this, we first describe existing normative frameworks, then point to functional description frameworks, and finally introduce concepts that we argue should be included in the bridging language. 

\begin{figure}[h]
    \centering
    \includegraphics[width=0.9\linewidth]{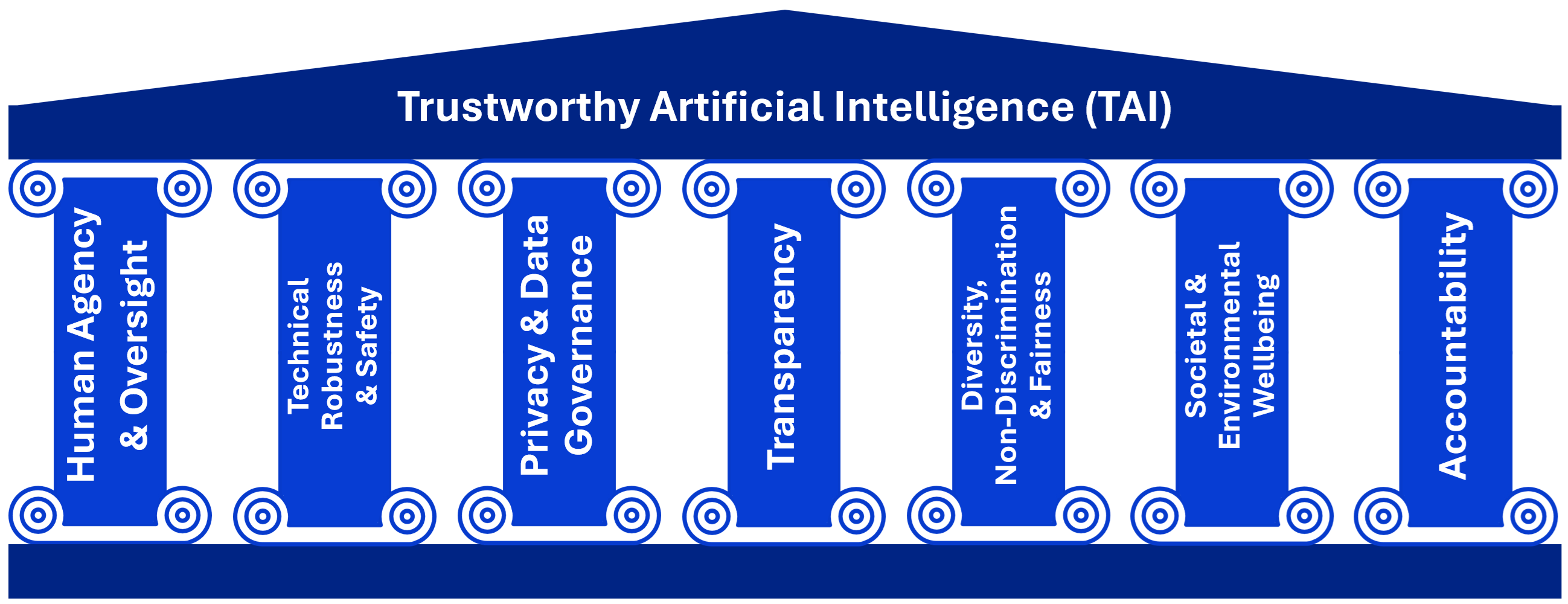}
    \caption{Seven normative key principles (pillars) of TAI as defined in \cite{hleg, aiactsection3}.}
    \label{fig:pillars}
\end{figure}

An important legal framework to consider is the \emph{AI Act}, which has formally gone into effect as of August 2024 as a legal framework proposed by the EU Commission \cite{union2021proposal, ai-guidelines-eu}. The AI Act aims to promote the adoption of TAI systems by taking a risk-based approach with rules for AI developers and deployers. They identify which systems are considered high-risk AI systems. The EU HLEG has defined seven key principles or pillars for Trustworthy AI \cite{hleg} as shown in Figure \ref{fig:pillars}, which have been introduced in the AI Act \cite{aiactsection3}. AI systems are expected to adhere to these trustworthiness principles, under continuous evaluation throughout their life cycle. These principles are a solid starting point, as they provide definitions and goals towards TAI. 

 Many parties have created guidelines and assessment tools for trustworthy AI. 
  The EU HLEG has created a list for self-assessment (ALTAI) \cite{altai}, OECD introduces guidelines \cite{oecd-ai-principles2019} with five values-based principles and recommendations to guide policymakers and AI actors. 
  NIST presents the AI Risk Management Framework (AIRMF) \cite{ai2023artificial}. 
  Several tools have been specifically developed to support evaluation in terms of the AI Act \cite{ai-act-compliance-checker, ai-act-implementation-tool, ai-impact-assessment-nl, altai}.
  These tools can serve as inspiration, though they focus primarily on risk identification and remain high-level. 
  
  Various semantic approaches have attempted to formalise Trustworthy AI \cite{golpayegani2022airo, keod22, newman2023taxonomy}, yet they lack explicit, unambiguous definitions on terminology like \triple{Transparency}. ALTAI provides a glossary for many terms related to Trustworthy AI, but these lack more rigorous semantics and still provide room for interpretation. 
  The aforementioned AIRO does capture terms like \triple{AISystem} and stakeholders (\triple{Provider}, \triple{Developer}, \triple{Deployer}) connected to norm sources. The Trustworthy Intelligent Systems Ontology (TISO) \cite{keod22} includes terms like \triple{Safety} and \triple{Transparency}, but these are not connected to any normative definitions. 
  Another approach is that of Lewis et al. \cite{Lewis21} who choose to model based on Activities, Entities and Agents, rather than characteristics such as transparency, as these are often not well-defined.

From the other perspective, multiple functional frameworks allow us to describe AI systems. To properly assess the trustworthiness of an AI system, one needs to know the system itself. It is therefore important to explore the design of the system to identify the key components and methods that are used within the system.
To ensure adoption, we recommend investigating techniques that are familiar to developers, such as UML-based modelling languages and extracting functional requirements from user stories \cite{ahmad2022requirementsengineeringartificialintelligence, lucassen2016improving, molla2024comparison, malan2001functional}.  For system descriptions, including system behaviour, formal semantics such as the System Ontology exists \cite{calhau2023exploring}. For a specific focus on AI systems, system modelling approaches such as Boxology \cite{BekkumBHMT21} are suitable because they lay the ground work for semantically defining AI components.

Closely related to what we envision is the Artificial Intelligence Trust Framework and Maturity Model (AI-TMM) \cite{mylrea2023ai}. 
This framework mentions evaluation options for trustworthiness properties. 
However, it does not provide a systematic approach for developers to explore their system.
Moreover, trustworthiness properties are not linked to existing normative frameworks.

\begin{figure}[h]
    \centering
    \includegraphics[width=0.99\linewidth]{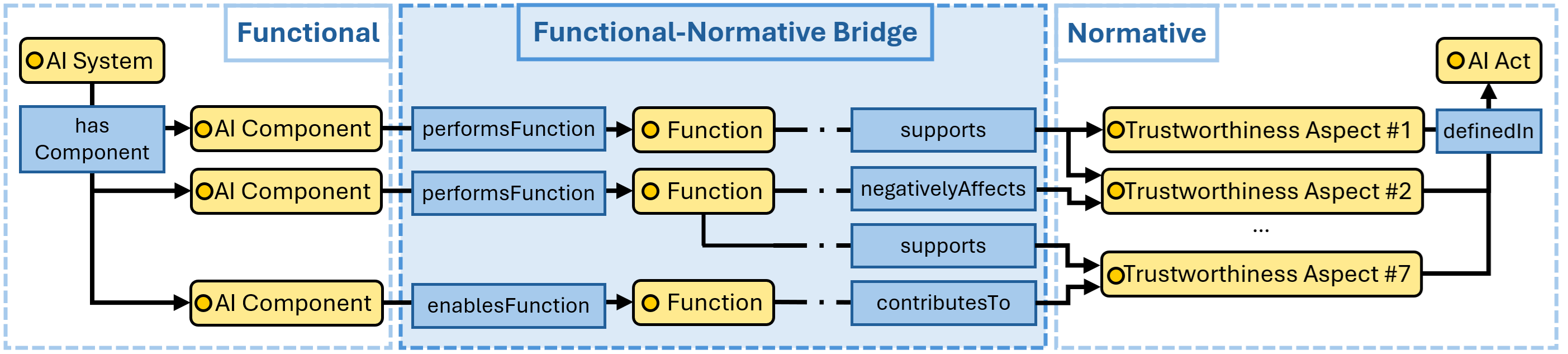}
    \caption{Conceptual language bridging functional aspects of AI components to normative (AI Act's) trustworthiness principles. Yellow blocks denote classes and blue labelled arrows denote relations.}
    \label{fig:bridge}
\end{figure}

We envision our approach for mapping from functional requirements to normative frameworks to be done in a way similar to what is described in Figure \ref{fig:bridge}. This shows an approach in which each AI system's component is described in terms of functionality ({\small \textsc{performsFunction Function}}), which is then mapped to trustworthiness aspects (TA) from various norms such as the AI Act's seven principles. For example, a recommender system can be used as AI component to perform recommendations (\triple{performsFunction}), which can support technical robustness (\triple{supports TA}\verb|#|1) by maintaining consistence, while it might risk being detrimental to bias (\triple{negativelyAffects TA}\verb|#|2) by over-relying on popular items leading to a lack of diversity. We believe that besides basing this framework on related frameworks, additional building blocks are required, threefold: 
1) The language should support a broad array of functional aspects. Instances of \triple{Function} could be \triple{ExtractData}, \triple{InferFacts}, \triple{SegmentText}, \triple{SearchSemantically}, \triple{SummarizeText}, \triple{PredictLabels}, \triple{DetectObject}, \triple{RetrieveRelatedKnowledge}, \triple{RankItems}, or \triple{ClassifyImage}. Having these functions predetermined helps developers uniformly describe AI systems.
2) As AI components do not operate in a vacuum, it is important for the language to describe the system on top of just its \triple{Function}s. Describing connections between AI components and from components to data is helpful. So, it is more insightful to include descriptions such has \triple{hasInput}, \triple{hasOutput}, \triple{dataFormat}, \triple{dataOrigin}, as well as different \triple{Function} relations such as \triple{evaluatedByFunction} or \triple{isDependentOnFunction}. 
3) Ample relations should be included to tie the functional descriptions to trustworthiness aspects. Not only positive relations (like \triple{supports}, \triple{contributesTo}) but also negative relations (e.g., \triple{negativelyAffects}) or even uncertain relations (e.g., \triple{possiblyAffects}). This to properly convey the possible impact of the AI system, without having to denote it just in terms of risk or positive influence.

\section{Discussion}

Finally, we outline several considerations and future actions for introducing a conceptual language to describe AI systems in terms of trustworthiness. One key consideration is that this language should not aim to provide a perfect delineation or definitive classification. It should not serve as a stamp of approval nor as a guarantee of correctness as such claims would be overly strong. Instead of being positioned as a self-assessment instrument, it should be an insight tool that supports and informs assessment processes and guides system development. Another important consideration is the inherent challenge of dealing with the dynamics of the gap we have identified. It can be difficult to exhaustively identify all AI system types and to create a fully comprehensive framework of functionalities as AI systems/functionality is an area that is constantly changing. Apart from technical developments, the field of trustworthiness is also continuously growing, requiring ongoing integration of new insights. 
Moreover, circular dependencies may occur as normative requirements may change based on technical limitations and vice versa. 

To deal with the ever-changing nature of the normative and functional requirements, we propose a targeted approach by e.g. initially focusing on specific types of AI systems and by focusing on a select number of pillars of the AI Act. This can then be updated and extended based on the users' needs.
This will ensure that the tool remains practical and manageable while still offering meaningful insights.

Looking ahead, we envision a scenario in which the conceptual language has been fully developed and embedded within a practical tool. Suppose a provider is tasked to describe their AI system, either as part of the development process or as a request from the deployer. Using the tool, they enter the system characteristics, after which the system automatically links to the underlying trust ontology. Based on this, the tool generates insights into the functional trustworthiness of the system in accordance with relevant standards and legislation. This helps guide the developer towards assessment of their systems' trustworthiness.

To advance towards this vision, future work needs to focus on several key directions. First, what we present here is an initial design and provides starting points from different points of view. To extend this, we expect to conduct a more comprehensive analysis of existing frameworks to identify appropriate building blocks for the conceptual language. Secondly, we intend to further develop and formalise the conceptual language in a standardised format (e.g., RDF). Finally, we would like to iteratively refine the framework based on validation and evaluation in real-world use cases.

Bridging this trustworthiness gap is crucial for ensuring that AI systems are not only functionally sound but also aligned with societal values and legal expectations. We believe that a positive impact can be made by helping stakeholders towards creating TAI according to standards, by providing understandable insights into the trustworthiness of their systems.

\section*{Declaration on Generative AI}

  The author(s) have not employed any Generative AI tools.

\bibliography{bibliography}

\end{document}